\pdfoutput=1

\documentclass[11pt]{article}

\usepackage[]{emnlp2021}

\usepackage{times}
\usepackage{latexsym}

\usepackage[T1]{fontenc}

\usepackage[utf8]{inputenc}

\usepackage{microtype}

\title{A Novel Corpus of Discourse Structure in Humans and Computers}

\author{Babak Hemmatian\textsuperscript{1,2}, \textbf{Sheridan Feucht}\textsuperscript{1}, \textbf{Rachel Avram}\textsuperscript{1}, \textbf{Alexander Wey}\textsuperscript{1}, \textbf{Muskaan Garg}\textsuperscript{1}, \textbf{Kate Spitalnic}\textsuperscript{3} \\
\textbf{Carsten Eickhoff}\textsuperscript{1}, \textbf{Ellie Pavlick}\textsuperscript{1}, \textbf{Bjorn Sandstede}\textsuperscript{1}, \textbf{Steven Sloman}\textsuperscript{1}\\

\textsuperscript{1}\textnormal{Brown University}, \textsuperscript{2}\textnormal{University of Illinois at Urbana-Champaign}, \textsuperscript{3}\textnormal{University of Sussex}\\
\texttt{babak.hemmatian@gmail.com}}

\begin{document}
\maketitle{}

\begin{abstract}
We present a novel corpus of 445 human- and computer-generated documents, comprising about 27,000 clauses, annotated for semantic clause types and coherence relations that allow for nuanced comparison of artificial and natural discourse modes. The corpus covers both formal and informal discourse, and contains documents generated using fine-tuned GPT-2 \citep{zellers2019} and GPT-3 \citep{brown2020}. We showcase the usefulness of this corpus for detailed discourse analysis of text generation by providing preliminary evidence that less numerous, shorter and more often incoherent clause relations are associated with lower perceived quality of computer-generated narratives and arguments.
\end{abstract}

\section{Introduction}

Recent years have seen the massive growth and popularity of text-generating algorithms, from GPT-2 \citep{radford2019} with its 1.5B parameters to GPT-3 with its 175B \citep{brown2020}. However, it is less clear which aspects of human discourse these models can or cannot capture. We present a novel corpus of human- and computer-generated text with detailed annotations of discourse elements and coherence to allow for more nuanced comparison of the two types of text.

\section{Corpus Composition}

Our corpus focuses on marijuana legalization discourse throughout 2008-2019, spanning a period of time throughout which general attitudes towards cannabis shifted, allowing our corpus to capture temporal changes in discourse style. The topical focus also allowed us to reduce sources of noise and focus instead on the discourse properties that distinguish human- and computer-produced content. The corpus contains 409 unique full-length documents, 445 unique annotations of them adding up to 26,986 clause labels, covering both formal news discourse and informal social media discussions. News articles from across the political spectrum (Washington Post and Huffington Post as Liberal and Fox News and Breitbart as Conservative sources) were sampled across the covered years from \href{www.commoncrawl.org}{Common Crawl}. Potentially relevant content was identified using \href{https://drive.google.com/drive/u/0/folders/1aBULVP4uYADJliPVQBRdg8IgwO\_YNsTU}{original comprehensive regular expressions} and then manually examined. Reddit discussions were carefully chosen from debate forums and other communities on Reddit (sampled from \href{https://github.com/BabakHemmatian/Marijuana\_Legalization\_Corpus\_Study}{this} comprehensive dataset).

This set of human-written documents is complemented by an almost equal number of corresponding computer-generated articles. 162 documents were generated using Grover \citep{zellers2019}, a fine-tuned version of GPT-2 \cite{radford2019} for news article generation using prompt text and meta-data. Another 60 annotated documents were produced using the most powerful GPT-3 engine once it was made available during the course of corpus development \citep{brown2020}. Each computer-generated document was produced using one of the human-written articles as a prompt, creating pairs of documents with similar content and style. Much trial and error was involved in prompt design to achieve the most coherent results. For GPT-3, various prompt lengths, top p, stop, and temperature parameter values were tested. For Grover, default parameters from the original work were eventually used \citep{zellers2019}, but we identified influences on the quality of generations that have not been recognized by the original authors. For instance, inclusion of article titles in the prompt (human or automatically-generated) was necessary for output coherence.


\section{Annotation Procedures}

Documents were annotated in terms of quality according to definitions of narrative and argument adapted from Smith \citeyearpar{smith2003}. Trained assistants (blind to whether documents were human-written or algorithmically-generated) rated each document for narrative and argument presence/absence. The ratings also included narrative quality across four dimensions (plausibility, completeness, consistency and coverage; \citealp{yale2013}) and argument quality across two dimensions (cogency and effectiveness; \citealp{wachsmuth2017}), as well as expressed attitude, partisanship, and other document-level information. Wachsmuth et al.’s third measure of argument quality was excluded after beginning annotation with all three measures, due to high correlation with cogency and effectiveness.

Only 57\% of documents generated by Grover contained the same amount of narrative-like discourse as their corresponding human-written prompts, whereas 68\% of Grover-generated documents maintained the presence or absence of argumentation in the prompt. The algorithm also lagged far behind humans when averaged across all quality measures (mean human quality = 4.5 (narrative), 4.41 (argument) on a 5-point scale; mean Grover quality = 2.275 (narrative), 1.915 (argument)). GPT-3 approached humans in performance but was still significantly worse across all measures of document quality (p < 0.01 across dimensions; mean quality = 4.01 (narrative), 3.54 (argument)).

To pinpoint what might be causing the disparity in document quality between humans and computers, the corpus was manually annotated by trained assistants for: 1) Structural linguistic elements of the two discourse modes based on the framework proposed by Smith \citeyearpar{smith2003} and developed for corpora by Friedrich \citeyearpar{friedrich2017}. Examples of clause types under this framework include basic states, bounded events and generic sentences. 2) A comprehensive set of coherence relations based on Wolf and Gibson \citeyearpar{wolfgibson2005}. Examples of relations between clauses in this framework include cause and effect, temporal sequence and contrast. Krippendorf’s alpha for interrater agreement ranged [.45,.52].

We extended these previous frameworks to better distinguish the compositional linguistic properties making up each clause label (e.g., based on \citealp{govindarajan2019}), and to account for incoherent content (e.g. repetition or intuitively meaningless relations) that may explain the difference in quality between computer and human discourse. More details about the annotation procedure and its evaluation can be found in Chapter 4 of \citealp{hemmatian2021}. The annotated corpus along with metadata and links to the code used in analyses can be found \href{https://github.com/sfeucht/annotation\_evaluation}{here}\footnote{https://github.com/sfeucht/annotation\_evaluation}. 

\section{Preliminary Results}

Ongoing analysis of the annotations reveals certain discourse aspects that the algorithms captured well, as well as other elements that differ wildly from their human counterparts. Topic distributions (based on LDA; \citealp{blei2003}) were largely similar between paired human and computer articles, suggesting that the algorithms can capture word co-occurrence patterns. The clause type compositions of narrative and argument discourse modes in artificial text also closely match those of human documents (\citealp{smith2003}; \citealp{friedrich2017}).

More differences were found between humans and computers in coherence relations. For certain types like cause-effect (exemplified by “because”), contrast (exemplified by “but”) and violated expectation (exemplified by “although”), more than a third of all Grover relations were incoherent. A correlation analysis between doc-level and clause-to-clause annotations showed that relations more commonly found in arguments showed a higher rate of incoherence than relations that were more frequent in narratives. The quality discrepancy was despite the fact that algorithms produced significantly fewer relations, particularly for the same categories, giving the model fewer chances to generated incoherent content. For instance, mean Grover document frequency of cause-effect relations was 1.71 (95 CI: [1.52,1.9]) compared with the human average of 3.56 (95 CI: [3.2,3.92]). For temporal sequence, a relation more commonly found in narratives, these means were much closer (Grover: 1.48 (95 CI: [1.35,1.61]); Human: 1.9 (95 CI: [1.71,2.08]). Computer-generated relations were also shorter in span across the board (p < 0.01), suggesting a less robust high-level document structure. GPT-3 showed a significant improvement over Grover, generating fewer incoherent relations, but the overall described patterns among coherence relation categories remained the same. For instance, cause-effect relations were less frequent than in human text (mean frequency of 1.5 compared with the human average of 3.56; p < 0.001), and up to half of instances for certain relation types like violated expectation (exemplified in the use of "although" or "but") were rated as incoherent.

These mechanisms may explain the quality discrepancy between humans and text-generation algorithms. However, more broadly speaking, they could reflect either training regimens employed for model development or the greater abstraction of certain (often more argument-related) relations that makes them inherently more difficult to capture. The results may also reflect how models trained on text completion tasks are not incentivized to learn what might be more often left implicit in human text, such as commonsense cause-effect relations \citep{becker2017}. More detailed analyses of discourse features that distinguish human- from computer-generated content, and subsequent disambiguation of these possibilities, requires further study. Therefore, we invite the computational discourse analysis community to aid us in further investigations of this novel corpus.


\section*{Acknowledgements}

This work is supported by a seed research grant from the Office of Vice President for Research at Brown University (GR300190). The current extended abstract was presented at the 2nd Workshop on Computational Approaches to Discourse, as part of the Empirical Methods in Natural Language Processing 2021 conference. We thank anonymous reviewers for their comments on its earlier version. We also thank Naomi Shammash for her work as an annotator and early research assistant for this project.

\bibliography{anthology,emnlp2021}
\bibliographystyle{aclnatbib}

\end{document}